\documentclass{article}
\usepackage[final]{conference}

\usepackage{microtype}
\usepackage{hyperref}
\usepackage{url}
\usepackage{booktabs}
\usepackage{amsmath,amsfonts,amssymb,bm}
\usepackage[pdftex]{graphicx}
\usepackage{lineno}

\definecolor{darkblue}{rgb}{0, 0, 0.5}
\hypersetup{colorlinks=true, citecolor=darkblue, linkcolor=darkblue, urlcolor=darkblue}

\usepackage{amsmath,amsfonts,bm}

\def\eqref#1{equation~\ref{#1}}

\def\1{\bm{1}}

\DeclareMathAlphabet{\mathsfit}{\encodingdefault}{\sfdefault}{m}{sl}
\SetMathAlphabet{\mathsfit}{bold}{\encodingdefault}{\sfdefault}{bx}{n}

\title{Structured Pruning of Large Language Models via Power Transformation and Sign-Preserving Score Aggregation with Adaptive Feature Retention}

\author{
\begin{tabular}{l}
\textbf{Ryota Kobayashi}$^{1}$, \textbf{Tsubasa Hirakawa}$^{1}$,
\textbf{Takayoshi Yamashita}$^{1}$, \textbf{Hironobu Fujiyoshi}$^{1}$, \\
\textbf{Yasunori Ishii}$^{2}$, \textbf{Tomoyuki Okuno}$^{2}$, \textbf{Kazuki Kozuka}$^{2}$ \\
$^{1}$Chubu University \\
$^{2}$Panasonic Holdings Corporation
\end{tabular}
}

\begin{document}

\ifcolmsubmission
\linenumbers
\fi

\maketitle

\begin{abstract}
This paper proposes an improved structured pruning method for large language models (LLMs) that addresses key challenges in adapting Adaptive Feature Retention (AFR), an unstructured pruning technique, to structured pruning. When applying AFR to structured pruning, three major problems arise: distribution mismatch between heterogeneous pruning scores, loss of sign information indicating optimization direction consistency, and influence of outliers. To address these issues, we propose a unified approach combining power transformation for nonlinear distribution alignment, sign-preserving score aggregation, and percentile-based outlier removal. Experiments on Llama-3-8B, Vicuna-v1.5-13B, and LLaVA-v1.5-13B demonstrate that our method maintains accuracy comparable to unstructured pruning while achieving practical inference speedup through structured pruning.
\end{abstract}

\section{Introduction}

Large language models (LLMs) demonstrate remarkable performance across diverse tasks, yet their billions of parameters impose substantial computational costs and memory requirements that hinder practical deployment. Pruning, which removes redundant weights, offers a promising approach to model compression.

Pruning methods are categorized into unstructured and structured pruning. Unstructured pruning removes individual weights independently, achieving high accuracy but providing limited practical speedup due to irregular sparsity patterns. Structured pruning removes entire neurons or channels, enabling efficient acceleration on standard hardware but often suffering from larger performance degradation due to reduced pruning granularity.

Adaptive Feature Retention (AFR)~\citep{afr} is an unstructured pruning method that combines feature-based ReFer~\citep{ReFer-L1} and gradient-based SNIP~\citep{snip} scores through standardization and summation, balancing preservation of pre-trained feature representations with adaptation to downstream tasks. When adapting AFR to structured pruning by aggregating weight-level scores to neuron-level scores, several challenges arise: distribution mismatch between heterogeneous scores (ReFer exhibits wide value ranges while SNIP concentrates in [0, 1]), loss of sign information indicating optimization direction consistency, and influence of outliers that dominate simple averaging.

To address these challenges, we propose an integrated approach combining power transformation for nonlinear distribution alignment, sign-preserving aggregation, and percentile-based outlier removal. Experiments on Llama-3-8B, Vicuna-v1.5-13B~\citep{vicuna}, and LLaVA-v1.5-13B~\citep{llava} demonstrate that our method significantly outperforms naive structured AFR and achieves comparable or superior performance to existing structured pruning methods while enabling 1.56-1.57$\times$ inference speedup at 50\% pruning rate.

Our main contributions are summarized as follows:
\begin{itemize}
\item We identify three fundamental challenges in adapting AFR to structured pruning: distribution mismatch between heterogeneous pruning scores, loss of sign information indicating optimization direction consistency, and influence of outliers during score aggregation.
\item We propose an integrated approach that addresses these challenges through (1) power transformation for nonlinear distribution alignment, (2) sign-preserving score aggregation to evaluate optimization direction consistency, and (3) percentile-based outlier removal for robust aggregation.
\item We conduct comprehensive experiments on three models (Llama-3-8B, Vicuna-v1.5-13B, and LLaVA-v1.5-13B) demonstrating that our method achieves substantial improvements over naive averaging (up to 21.27 points) and outperforms existing structured pruning methods while enabling practical inference speedup (1.56-1.57$\times$ at 50\% pruning).
\end{itemize}

\section{Background: Adaptive Feature Retention}

Adaptive Feature Retention (AFR)~\citep{afr} is an unstructured pruning method for pre-trained models that combines ReFer~\citep{ReFer-L1} and SNIP~\citep{snip} to balance preservation of learned feature spaces with adaptation to downstream tasks. The AFR pruning score $S_{AFR}(\theta_n)$ for weight $\theta_n$ is defined as:
\begin{equation}
    S_{AFR}(\theta_n) = \mathcal{Z}\left( \left|\frac{\partial \sum_{l=1}^{L} F^l_{svd}}{\partial \theta_n} \theta_n\right|\right) + \mathcal{Z}\left(\left| \frac{\partial \mathcal{L}}{\partial \theta_n} \theta_n \right|\right)
    \label{eq:afr}
\end{equation}
where $\mathcal{Z}(\cdot)$ denotes standardization (z-score normalization), the first term represents ReFer, and the second term represents SNIP. Here, $F^l_{svd}$ is the mean singular value from singular value decomposition of layer $l$'s output features, and $\mathcal{L}$ is the task-specific loss function.

ReFer focuses on maintaining feature representations by measuring how each weight affects the feature space geometry captured during pre-training. It computes the sensitivity of feature space characteristics with respect to each weight, identifying parameters that are crucial for preserving the learned representations. However, while ReFer excels at maintaining pre-trained knowledge, it may inadvertently remove weights that are necessary for adapting to downstream tasks, as it does not directly consider task-specific objectives.

SNIP, on the other hand, uses the product of loss gradients and weights as the pruning score, directly measuring each weight's contribution to the task-specific loss function. This gradient-based approach naturally identifies parameters important for the current task. However, for pre-trained models, many weights have already converged to near-optimal values, resulting in small gradients ($\frac{\partial \mathcal{L}}{\partial \theta_n} \approx 0$) that do not accurately reflect their importance. This makes SNIP alone insufficient for proper importance evaluation in the context of pre-trained models.

By combining both scores through standardization and summation, AFR achieves effective pruning that preserves pre-trained knowledge while enabling task adaptation. The standardization ensures that both scores contribute equally despite their different scales, allowing the combined score to capture both feature preservation and task relevance.

\section{Challenges in adapting AFR to structured pruning}

To enable inference acceleration with AFR, we apply its weight-level pruning scores to structured pruning. This section identifies the problems that arise in this adaptation.

\subsection{Naive aggregation for structured pruning}

\begin{figure}[t]
    \centering
    \includegraphics[width=\linewidth]{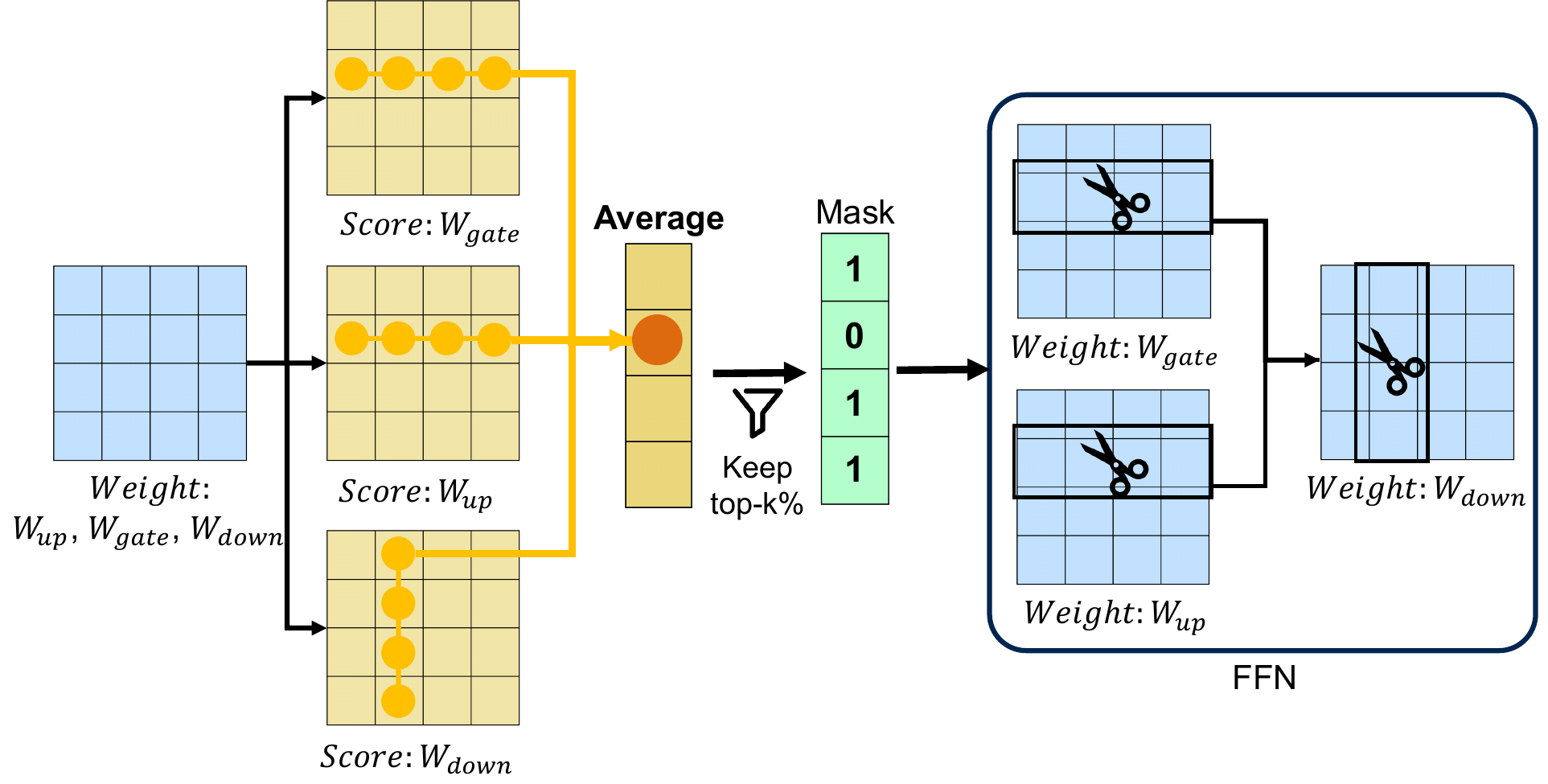}
    \caption{Pruning score aggregation process for structured pruning.}
    \label{fig:network}
\end{figure}

To adapt the unstructured AFR method to structured pruning, we consider naive averaging for score aggregation as illustrated in Figure~\ref{fig:network}. First, we compute pruning scores for each element of the weight matrix. Then, we aggregate scores within each neuron and compute the average score $\bar{S}_j$ at the neuron level as:
\begin{equation}
    \bar{S}_j = \frac{1}{m} \sum_{i=1}^{m} |S_{ij}|
    \label{eq:naive}
\end{equation}
where $S_{ij}$ is the pruning score of the $i$-th weight in neuron $j$, and $m$ is the number of weights per neuron. Finally, neurons with the lowest average scores are removed according to the target pruning rate.

\begin{table}[t]
\centering
\small
\caption{Preliminary experimental results (accuracy \%).}
\label{tab:preliminary}
\begin{tabular}{@{}llccccc@{}}
\toprule
\textbf{Pruning} & \textbf{Method} & \textbf{WinoG} & \textbf{HellaS} & \textbf{ARC-e} & \textbf{ARC-c} & \textbf{MMLU} \\
\midrule
0\% & Llama-3-8B & 72.61 & 79.16 & 77.74 & 53.33 & 62.13 \\
\midrule
20\% & AFR (unstruct.) & 71.59 & 73.79 & 76.94 & 48.38 & 58.85 \\
& AFR (naive avg.) & 59.35 & 53.63 & 43.48 & 29.35 & 30.14 \\
\midrule
50\% & AFR (unstruct.) & 60.62 & 50.64 & 55.35 & 32.94 & 35.07 \\
& AFR (naive avg.) & 52.25 & 29.29 & 29.92 & 26.02 & 23.07 \\
\bottomrule
\end{tabular}
\end{table}

\subsection{Problems with naive aggregation}

Table~\ref{tab:preliminary} shows that applying naive averaging to AFR on LLMs results in severe performance degradation on natural language benchmarks. We identify three underlying problems:

\textbf{Problem 1: Distribution mismatch between heterogeneous scores.} ReFer and SNIP scores have fundamentally different distribution characteristics. ReFer scores span a wide value range with relatively uniform distribution, while SNIP scores concentrate in the narrow [0, 1] range due to small gradients in pre-trained models. Simple standardization in Eq.~\eqref{eq:afr} linearly amplifies the minute differences among SNIP's densely clustered low-importance parameters, introducing noise into the AFR scores. This noise disrupts ReFer's importance judgments and degrades pruning accuracy.

\textbf{Problem 2: Loss of optimization direction consistency.} ReFer and SNIP scores are computed as products of gradients and weights, where the sign encodes weight behavior during optimization. In the weight update rule $\theta \leftarrow \theta - \eta \frac{\partial L}{\partial \theta}$, positive scores indicate weights decreasing in magnitude, while negative scores indicate increasing magnitude. For structured pruning, the consistency of optimization directions within a neuron is critical. When all weights in a neuron optimize in the same direction, the neuron has a structurally coherent role. Conversely, mixed directions indicate internal cancellation effects and limited overall contribution. However, Eq.~\eqref{eq:naive} takes absolute values before averaging, discarding this directional consistency information and preventing proper neuron-level importance evaluation.

\begin{figure}[t]
    \centering
    \includegraphics[width=0.9\columnwidth]{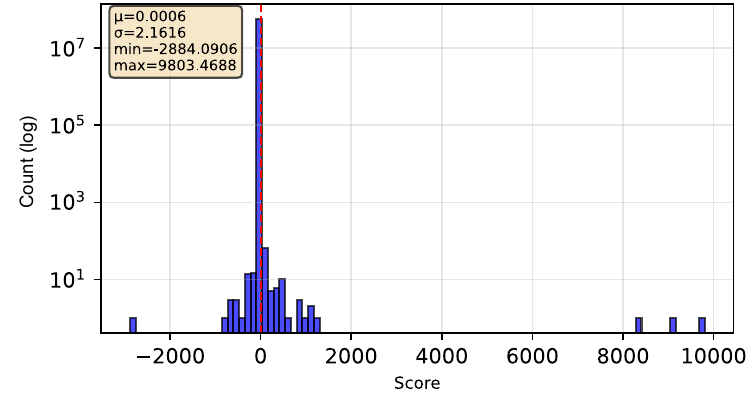}
    \caption{Example distribution of AFR pruning scores showing concentration in the center with extreme outliers at the tails.}
    \label{fig:fo_score}
\end{figure}

\textbf{Problem 3: Influence of outliers.} Figure~\ref{fig:fo_score} shows the weight-level score distribution for ReFer in one layer. Most scores concentrate in a relatively narrow range, but extreme outliers exist at both tails. The full score range spans $-2{,}884$ to $9{,}803$ (width: $12{,}687$), but removing the extreme 2\% reduces the range width to $48.8$. With naive averaging, the presence or absence of outliers can reverse the relative ordering of neuron-level scores, preventing proper importance evaluation.

\section{Proposed method}

We propose an integrated approach combining three techniques to address the problems identified above.

\subsection{Nonlinear distribution alignment via power transformation}
\label{sec:power}

To address Problem 1, we apply power transformation to SNIP scores before standardization. For SNIP score $s_n^{\text{SNIP}}$, we apply:
\begin{equation}
\hat{s}_n^{\text{SNIP}} = \left(s_n^{\text{SNIP}}\right)^b
\label{eq:power}
\end{equation}
where $b > 1$. For $0 \leq s_n^{\text{SNIP}} < 1$, this transformation has two key properties that address the distribution mismatch problem.

First, it provides noise suppression in the low-importance region. Since $x^b \ll x$ for values near 0, minute differences in low-importance parameters are nonlinearly compressed, effectively filtering out noise from the densely clustered low-importance region of SNIP scores.

Second, it enables signal preservation in the high-importance region. For values near 1, $x^b \approx x$ holds approximately, preserving information for high-importance parameters. This asymmetric behavior creates a soft thresholding effect, where noise suppression strength adjusts continuously according to importance magnitude.

\begin{figure}[t]
\centering
\includegraphics[width=\linewidth]{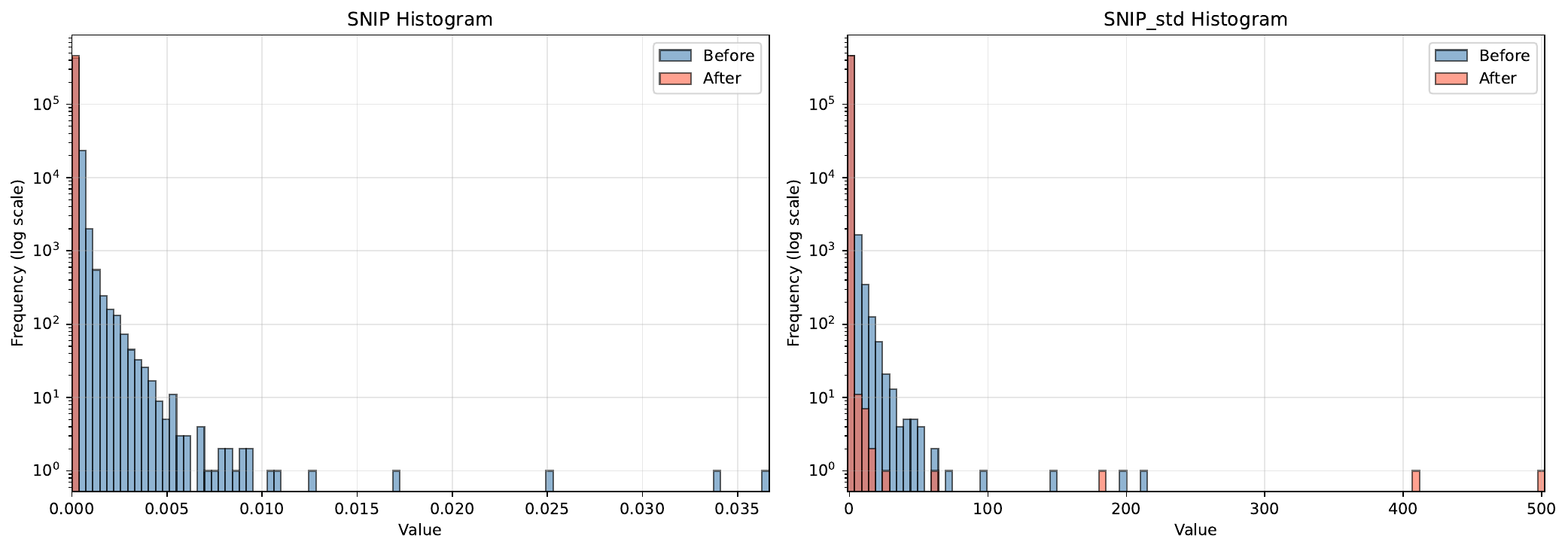}
\caption{Distribution comparison before and after power transformation ($b=2.7$). Left: SNIP score distributions before standardization show how power transformation compresses the densely clustered low-value region. Right: Distributions after standardization demonstrate improved concentration, with the transformed distribution (orange) exhibiting reduced spread in the low-importance region while maintaining separation of high-importance values.}
\label{fig:power_histogram}
\end{figure}

After power transformation, we apply standardization to the transformed scores. To quantify the improvement in distribution quality, we analyze the signal-to-noise ratio (SNR). Let $\mu_{\text{signal}}$ denote the mean absolute value of high-importance scores (top 10\% by magnitude) and $\sigma_{\text{noise}}$ denote the standard deviation of low-importance scores (bottom 50\%). The SNR before and after power transformation can be expressed as:
\begin{equation}
\text{SNR}_{\text{original}} = \frac{\mu_{\text{signal}}}{\sigma_{\text{noise}}}, \quad
\text{SNR}_{\text{transformed}} = \frac{\mu_{\text{signal}}^b}{\sigma_{\text{noise}}^b}
\label{eq:snr}
\end{equation}
Since SNIP scores concentrate in [0, 1] with most values near 0, we have $\sigma_{\text{noise}} < \mu_{\text{signal}} < 1$. For $b > 1$, the power function satisfies $x^b \ll x$ for $x \ll 1$ and $x^b \lesssim x$ for $x \approx 1$, which means $\sigma_{\text{noise}}^b \ll \mu_{\text{signal}}^b$. Therefore, the ratio $\text{SNR}_{\text{transformed}}/\text{SNR}_{\text{original}} = (\mu_{\text{signal}}/\sigma_{\text{noise}})^{b-1}$ increases exponentially with $b$, demonstrating that power transformation amplifies the relative distinction between signal and noise. Figure~\ref{fig:power_histogram} illustrates this effect: the left histogram shows how power transformation compresses the low-value region before standardization, while the right histogram demonstrates how this compression leads to improved concentration after standardization.

The power transformation reduces the variance $\sigma_{\hat{S}}^2$ compared to the original scores, causing the standardization division to have a stronger stretching effect on the remaining distribution. However, critically, only the signal component remaining after transformation is stretched, while noise has already been suppressed in the preceding step. This improved SNR during scale adjustment enables more effective integration with ReFer scores.

\subsection{Sign-preserving score aggregation}
\label{sec:abs}

To address Problem 2, we propose aggregation with post-averaging absolute values. Instead of taking absolute values before averaging as in Eq.~\eqref{eq:naive}, we average the signed pruning scores within each neuron, then take the absolute value after averaging.

This approach leverages sign information to evaluate optimization direction consistency. If all weights have consistent signs, the average retains a large magnitude. Conversely, if weights have mixed signs, the average approaches zero due to cancellation. This naturally quantifies the degree of directional consistency.

Neurons with consistent optimization directions indicate structurally coherent roles where all weights cooperate toward the same objective, making them valuable for preservation. In contrast, neurons with mixed directions exhibit internal cancellation effects, suggesting limited overall contribution. By preserving sign information during aggregation, our method enables proper evaluation of this structural property for structured pruning.

\subsection{Percentile-based outlier removal}
\label{sec:outlier}

To mitigate Problem 3's outlier influence, we introduce percentile-based outlier removal. To determine the optimal removal rate, we analyzed pruning score distribution statistics (Table~\ref{tab:outlier_stats}).

At 2\% removal, the range decreases to 48.8, and kurtosis drops from 16,488,324 to 3,505 (a 99.98\% reduction). Kurtosis quantifies distribution tail heaviness; the original distribution's extremely high kurtosis indicates a long-tailed distribution. From 1\% to 2\% removal, kurtosis decreases by 94.2\%, but from 2\% to 3\%, the decrease is only 54.9\%, suggesting outlier removal effects saturate beyond 2\%.

The interquartile range (IQR) represents the spread of the central 50\% of the distribution. Substantial IQR changes indicate removal of not just outliers but also central distribution information, suggesting information loss. The original IQR is 91.72, and at 2\% removal it remains 86.71, preserving the structure, while at 10\% removal it becomes 68.20.

These results confirm that 2\% removal optimally eliminates outlier influence (from range and kurtosis perspectives) while preserving original distribution information (from IQR perspective).

\begin{table}[t]
\centering
\small
\caption{Statistical changes by outlier removal rate.}
\label{tab:outlier_stats}
\begin{tabular}{@{}ccccc@{}}
\toprule
\textbf{Removal} & \textbf{Range} & \textbf{Range} & \textbf{Kurtosis} & \textbf{IQR} \\
\textbf{rate} & & \textbf{reduction} & & ($\times 10^3$) \\
\midrule
0\% & 12,687 & --- & 16,488,324 & 91.72 \\
1\% & 161.4 & 98.7\% & 60,133.04 & 89.23 \\
2\% & 48.8 & 99.6\% & 3,504.79 & 86.71 \\
3\% & 27.6 & 99.8\% & 1,581.82 & 84.25 \\
5\% & 17.6 & 99.9\% & 891.48 & 79.46 \\
10\% & 9.42 & 99.9\% & 544.83 & 68.20 \\
\bottomrule
\end{tabular}
\end{table}

Therefore, we compute the 2nd percentile $P_2$ and 98th percentile $P_{98}$ for each neuron's score set and remove scores outside these thresholds.

\subsection{Integrated approach}
\label{sec:integration}

We combine the techniques from Sections~\ref{sec:power}--\ref{sec:outlier} into a unified method:

\begin{enumerate}
\item \textbf{Outlier removal.} Remove scores outside the 2nd and 98th percentiles for each neuron's ReFer and SNIP score sets.

\item \textbf{Sign-preserving aggregation.} Compute the mean of cleaned scores while maintaining signs, then take the absolute value to obtain neuron-level scores.

\item \textbf{Power transformation.} Apply power transformation (Eq.~\eqref{eq:power}) to aggregated neuron-level SNIP scores.

\item \textbf{Score combination.} Sum transformed neuron-level SNIP scores with neuron-level ReFer scores to obtain final pruning scores.
\end{enumerate}

Concretely, for each neuron, we first remove weight-level scores outside the 2nd and 98th percentiles for both ReFer and SNIP. We then compute the average of the remaining scores while preserving signs, followed by taking absolute values to obtain neuron-level scores for ReFer and SNIP separately. The power transformation with exponent $b$ is applied to the neuron-level SNIP score. Finally, we sum the transformed SNIP score with the ReFer score to obtain the final neuron-level pruning score.

This integrated approach sequentially applies outlier suppression, sign consistency evaluation, and heterogeneous score distribution alignment. Each component addresses a specific problem while building upon the outputs of previous steps, enabling stable and accurate pruning score evaluation for structured pruning.

\section{Experiments}

We evaluate our proposed method on Llama-3-8B, Vicuna-v1.5-13B, and LLaVA-v1.5-13B.

\subsection{Experimental setup}

We apply structured pruning to the feed-forward networks (FFNs) in each block of the models. FFNs constitute a substantial portion of model parameters and are amenable to structured pruning due to their layer-wise independence. For Llama-3-8B and Vicuna-v1.5-13B, which are language-only models, we apply our method to all FFN layers across the transformer blocks. Each FFN consists of multiple linear layers that can be pruned at the neuron level.

LLaVA-v1.5-13B presents a more complex scenario as it consists of a language part (Vicuna) and vision part (CLIP) connected through projection layers. For the language part where calibration data has ground-truth labels, we apply the full AFR method (combining ReFer and SNIP) to prune FFNs. For the vision part, CLIP functions as a feature extractor without direct downstream task supervision, making SNIP's loss gradients undefined since there is no task-specific loss to differentiate through the vision encoder. Therefore, we prune CLIP's fully connected layers using only the ReFer component, which relies solely on feature space geometry and does not require task-specific gradients.

We evaluate at pruning rates of 20\% and 50\%, widely adopted in prior work as representative of moderate and aggressive pruning scenarios. The power transformation exponent is set to $b=2.7$ based on systematic ablation experiments described in Appendix~\ref{sec:power_appendix}, which demonstrates that this value achieves optimal performance across multiple benchmarks. For Llama-3-8B and Vicuna-v1.5-13B, we use five benchmark datasets covering diverse language understanding capabilities: WinoGrande~\citep{winogrande} for commonsense reasoning, HellaSwag~\citep{hellaswag} for commonsense natural language inference, ARC-easy and ARC-challenge~\citep{arc} for question answering, and MMLU~\citep{mmlu} for multitask language understanding. For LLaVA-v1.5-13B, we use three vision-language benchmarks: GQA~\citep{gqa} for visual reasoning, VizWiz~\citep{vizwiz} for visual question answering, and ScienceQA~\citep{scienceqa} for multimodal science questions. Accuracy serves as the evaluation metric across all benchmarks. We compare against unstructured AFR, structured AFR with naive averaging, and existing structured pruning methods including LLM-Pruner~\citep{llmpruner}, LoRAP~\citep{lorap}, and CFSP~\citep{cfsp}.

\begin{table}[t]
\centering
\small
\caption{Comparison with existing methods on Llama-3-8B (accuracy \%).}
\label{tab:exp_comparison_llama}
\begin{tabular}{@{}llccccc@{}}
\toprule
\textbf{Pruning} & \textbf{Method} & \textbf{WinoG} & \textbf{HellaS} & \textbf{ARC-e} & \textbf{ARC-c} & \textbf{MMLU} \\
\midrule
0\% & Llama-3-8B & 72.61 & 79.16 & 77.74 & 53.33 & 62.13 \\
\midrule
\multicolumn{7}{@{}l@{}}{\textit{20\% pruning}} \\
& AFR (unstruct.) & 71.59 & 73.79 & 76.94 & 48.38 & 58.85 \\
& AFR (naive avg.) & 59.35 & 53.63 & 43.48 & 29.35 & 30.14 \\
& AFR (proposed) & \textbf{74.03} & \textbf{72.62} & \textbf{68.86} & \textbf{45.73} & \textbf{61.05} \\
& LLM-Pruner & 69.85 & 69.02 & 63.59 & 40.53 & 48.36 \\
& LoRAP & 71.19 & 70.48 & 69.61 & 45.48 & 44.57 \\
& CFSP & 68.67 & 68.06 & 67.63 & 42.92 & 50.71 \\
\midrule
\multicolumn{7}{@{}l@{}}{\textit{50\% pruning}} \\
& AFR (unstruct.) & 60.62 & 50.64 & 55.35 & 32.94 & 35.07 \\
& AFR (naive avg.) & 52.25 & 29.29 & 29.92 & 26.02 & 23.07 \\
& AFR (proposed) & \textbf{61.80} & \textbf{46.38} & 45.20 & \textbf{30.29} & \textbf{52.51} \\
& LLM-Pruner & 52.49 & 35.50 & 37.88 & 25.51 & 22.95 \\
& LoRAP & 57.30 & 40.19 & 42.60 & 26.79 & 26.85 \\
& CFSP & 57.06 & 43.13 & \textbf{48.53} & 28.50 & 26.61 \\
\bottomrule
\end{tabular}
\end{table}

\subsection{Comparison with existing methods}

Tables~\ref{tab:exp_comparison_llama}, \ref{tab:exp_comparison_vicuna}, and \ref{tab:exp_comparison_llava} present results for Llama-3-8B, Vicuna-v1.5-13B, and LLaVA-v1.5-13B, respectively.

For Llama-3-8B, our method achieves 64.46 at 20\% pruning and 47.24 at 50\% pruning, representing improvements of 21.27 and 15.13 points over naive averaging. Compared to unstructured AFR, our method shows only 1.45 point difference at 20\% and exceeds it by 0.32 points at 50\%, maintaining performance comparable to unstructured pruning while enabling structured acceleration. Compared to existing structured methods, our method outperforms the average by 3.60 points at 20\% and 9.04 points at 50\%. At 50\% pruning, the improvements are substantial: 12.37 points over LLM-Pruner, 8.49 points over LoRAP, and 6.47 points over CFSP.

Similar trends appear for Vicuna-v1.5-13B. Our method achieves 63.70\% at 20\% and 52.91\% at 50\%, exceeding existing method averages by 1.76 and 6.40 points respectively. At 50\% pruning, our method maintains a clear advantage, with CFSP being the closest competitor at 51.20\% average accuracy.

For the vision-language model LLaVA-v1.5-13B, our method achieves 62.83\% at 20\% and 47.08\% at 50\%, demonstrating effectiveness across different model architectures including VLMs. At 20\% pruning, performance slightly exceeds the original model by 0.06 points, indicating effective identification and removal of redundant parameters.

The performance gains result from the synergistic combination of distribution alignment, sign consistency evaluation, and outlier suppression. The power transformation addresses heterogeneous score distributions, sign-preserving aggregation captures structural properties of neuron-level optimization, and outlier removal ensures stable aggregation resistant to extreme values.

\begin{table}[tb]
\centering
\small
\caption{Comparison with existing methods on Vicuna-v1.5-13B (accuracy \%).}
\label{tab:exp_comparison_vicuna}
\begin{tabular}{@{}llccccc@{}}
\toprule
\textbf{Pruning} & \textbf{Method} & \textbf{WinoG} & \textbf{HellaS} & \textbf{ARC-e} & \textbf{ARC-c} & \textbf{MMLU} \\
\midrule
0\% & Vicuna-v1.5-13B & 72.51 & 77.46 & 74.83 & 50.68 & 54.50 \\
\midrule
20\% & Proposed & \textbf{69.61} & 73.91 & 72.94 & \textbf{48.29} & \textbf{54.03} \\
& LLM-Pruner & 67.88 & 74.16 & \textbf{73.30} & 46.93 & 41.72 \\
& LoRAP & 68.90 & \textbf{75.15} & 71.55 & 44.97 & 49.88 \\
& CFSP & 68.82 & 73.86 & 71.21 & 45.14 & 50.93 \\
\midrule
50\% & Proposed & \textbf{64.48} & 53.82 & \textbf{57.53} & \textbf{37.97} & \textbf{50.74} \\
& LLM-Pruner & 58.48 & 42.23 & 57.45 & 32.08 & 23.00 \\
& LoRAP & 60.46 & 43.81 & 55.89 & 33.62 & 29.59 \\
& CFSP & 62.04 & \textbf{58.05} & 57.20 & 37.71 & 40.99 \\
\bottomrule
\end{tabular}
\end{table}

\begin{table}[tb]
\centering
\small
\caption{Evaluation results on LLaVA-v1.5-13B (accuracy \%).}
\label{tab:exp_comparison_llava}
\begin{tabular}{@{}llccc@{}}
\toprule
\textbf{Pruning} & \textbf{Method} & \textbf{GQA} & \textbf{VizWiz} & \textbf{ScienceQA} \\
\midrule
0\% & LLaVA-v1.5-13B & 63.29 & 56.15 & 72.83 \\
20\% & Proposed & \textbf{60.34} & \textbf{58.28} & \textbf{70.05} \\
50\% & Proposed & \textbf{29.19} & \textbf{48.50} & \textbf{63.56} \\
\bottomrule
\end{tabular}
\end{table}

\subsection{Resource reduction evaluation}

Tables~\ref{tab:resource_reduction_llama} and \ref{tab:resource_reduction_llava} show resource reduction effects for each pruning rate. For Llama-3-8B, 20\% pruning achieves 14.0\% parameter reduction (from 8.03B to 6.90B), 14.0\% VRAM reduction (from 16.06GB to 13.08GB), and 1.10$\times$ inference speedup. At 50\% pruning, the reductions become more substantial: 35.1\% parameter reduction (to 5.21B), 35.1\% VRAM reduction (to 10.42GB), and 1.57$\times$ inference speedup.

For LLaVA-v1.5-13B, 20\% pruning achieves 13.0\% parameter reduction (from 13.3B to 11.4B), 11.7\% VRAM reduction, and 1.20$\times$ speedup. At 50\% pruning, LLaVA-v1.5-13B achieves 32.6\% parameter reduction (to 8.98B), 24.9\% VRAM reduction (to 25.55GB), and 1.56$\times$ speedup.

The speedup is less than the theoretical maximum (2$\times$ for 50\% pruning) because (1) only FFN layers are pruned while attention layers remain intact, and (2) memory access costs and kernel launch overhead become more significant proportionally. Nevertheless, the achieved speedups represent practical improvements for deployment scenarios. Structured pruning enables this acceleration without specialized sparse computation libraries, allowing the pruned models to run efficiently on standard hardware using dense matrix operations.

\begin{table}[tb]
\centering
\small
\caption{Resource reduction evaluation for Llama-3-8B.}
\label{tab:resource_reduction_llama}
\begin{tabular}{@{}lccccc@{}}
\toprule
\textbf{Pruning} & \multicolumn{2}{c}{\textbf{Parameters}} & \multicolumn{2}{c}{\textbf{VRAM (GB)}} & \textbf{Speed (ms/sample)} \\
\midrule
0\% & 8.03B & --- & 16.06 & --- & 22.98 (1.0$\times$) \\
20\% & 6.90B & 14.0\%$\downarrow$ & 13.08 & 14.0\%$\downarrow$ & 20.89 (1.10$\times$) \\
50\% & 5.21B & 35.1\%$\downarrow$ & 10.42 & 35.1\%$\downarrow$ & 14.64 (1.57$\times$) \\
\bottomrule
\end{tabular}
\end{table}

\begin{table}[t]
\centering
\small
\caption{Resource reduction evaluation for LLaVA-v1.5-13B.}
\label{tab:resource_reduction_llava}
\begin{tabular}{@{}lccccc@{}}
\toprule
\textbf{Pruning} & \multicolumn{2}{c}{\textbf{Parameters}} & \multicolumn{2}{c}{\textbf{VRAM (GB)}} & \textbf{Speed (ms/sample)} \\
\midrule
0\% & 13.3B & --- & 34.04 & --- & 756.1 (1.0$\times$) \\
20\% & 11.4B & 13.0\%$\downarrow$ & 30.05 & 11.7\%$\downarrow$ & 630.6 (1.20$\times$) \\
50\% & 8.98B & 32.6\%$\downarrow$ & 25.55 & 24.9\%$\downarrow$ & 481.9 (1.56$\times$) \\
\bottomrule
\end{tabular}
\end{table}

\section{Conclusion}

This paper proposes an improved structured pruning method that maintains accuracy while enabling inference acceleration. We identified three key challenges in adapting AFR to structured pruning: distribution mismatch between heterogeneous scores, loss of sign information, and outlier influence. To address these, we propose an integrated approach combining power transformation for nonlinear distribution alignment, sign-preserving aggregation, and percentile-based outlier removal.

Experiments on Llama-3-8B, Vicuna-v1.5-13B, and LLaVA-v1.5-13B demonstrated the effectiveness of our approach. Our method achieves substantial improvements over naive averaging (21.27 and 15.13 points at 20\% and 50\% pruning) and outperforms existing structured pruning methods. At 50\% pruning on Llama-3-8B, our method achieves 35.1\% parameter reduction and 1.57$\times$ inference speedup. For LLaVA-v1.5-13B, we achieve 32.6\% parameter reduction and 1.56$\times$ speedup. Ablation studies confirm synergistic effects of all proposed components.

Future work includes extending to attention modules, optimizing layer-wise adaptive pruning rates.

\newpage

\bibliography{paper}

@inproceedings{afr,
  title={Single-shot Foresight Pruning Balancing Knowledge Retention and Downstream Task Adaptation for Pre-trained Models},
  author={Nitta, Tsunemi and Hirakawa, Tsubasa and Yamashita, Takayoshi and Fujiyoshi, Hironobu},
  booktitle={Meeting on Image Recognition and Understanding (MIRU)},
  year={2025}
}

@inproceedings{ReFer-L1,
  title={Single-shot Foresight Pruning for Maintaining Feature Representations of Pre-trained Models},
  author={Nitta, Tsunemi and Kohama, Hirokazu and Hirakawa, Tsubasa and Yamashita, Takayoshi and Fujiyoshi, Hironobu},
  booktitle={Meeting on Image Recognition and Understanding (MIRU)},
  year={2024}
}

@inproceedings{snip,
  title={{SNIP}: Single-shot Network Pruning based on Connection Sensitivity},
  author={Lee, Namhoon and Ajanthan, Thalaiyasingam and Torr, Philip H. S.},
  booktitle={International Conference on Learning Representations (ICLR)},
  year={2019}
}

@inproceedings{vicuna,
  title={Judging {LLM}-as-a-Judge with {MT-Bench} and {Chatbot Arena}},
  author={Zheng, Lianmin and Chiang, Wei-Lin and Sheng, Ying and Zhuang, Siyuan and Wu, Zhanghao and Zhuang, Yonghao and Lin, Zi and Li, Zhuohan and Li, Dacheng and Xing, Eric P. and Zhang, Hao and Gonzalez, Joseph E. and Stoica, Ion},
  booktitle={Advances in Neural Information Processing Systems (NeurIPS)},
  year={2023}
}

@inproceedings{llava,
  title={Improved Baselines with Visual Instruction Tuning},
  author={Liu, Haotian and Li, Chunyuan and Li, Yuheng and Lee, Yong Jae},
  booktitle={IEEE/CVF Conference on Computer Vision and Pattern Recognition (CVPR)},
  pages={26286--26296},
  year={2024},
  doi={10.1109/CVPR52733.2024.02484}
}

@inproceedings{winogrande,
  title={{WinoGrande}: An Adversarial {Winograd} Schema Challenge at Scale},
  author={Sakaguchi, Keisuke and Le Bras, Ronan and Bhagavatula, Chandra and Choi, Yejin},
  booktitle={AAAI Conference on Artificial Intelligence (AAAI)},
  year={2020}
}

@inproceedings{hellaswag,
  title={{HellaSwag}: Can a Machine Really Finish Your Sentence?},
  author={Zellers, Rowan and Holtzman, Ari and Bisk, Yonatan and Farhadi, Ali and Choi, Yejin},
  booktitle={Annual Meeting of the Association for Computational Linguistics (ACL)},
  year={2019}
}

@article{arc,
  title={Think you have Solved Question Answering? Try {ARC}, the {AI2} Reasoning Challenge},
  author={Clark, Peter and Cowhey, Isaac and Etzioni, Oren and Khot, Tushar and Sabharwal, Ashish and Schoenick, Carissa and Tafjord, Oyvind},
  journal={arXiv preprint arXiv:1803.05457},
  year={2018}
}

@inproceedings{mmlu,
  title={Measuring Massive Multitask Language Understanding},
  author={Hendrycks, Dan and Burns, Collin and Basart, Steven and Zou, Andy and Mazeika, Mantas and Song, Dawn and Steinhardt, Jacob},
  booktitle={International Conference on Learning Representations (ICLR)},
  year={2021}
}

@inproceedings{gqa,
  title={{GQA}: A New Dataset for Real-World Visual Reasoning and Compositional Question Answering},
  author={Hudson, Drew A. and Manning, Christopher D.},
  booktitle={IEEE/CVF Conference on Computer Vision and Pattern Recognition (CVPR)},
  year={2019}
}

@inproceedings{vizwiz,
  title={{VizWiz} Grand Challenge: Answering Visual Questions from Blind People},
  author={Gurari, Danna and Li, Qing and Stangl, Abigale J. and Guo, Anhong and Lin, Chi and Grauman, Kristen and Luo, Jiebo and Bigham, Jeffrey P.},
  booktitle={IEEE/CVF Conference on Computer Vision and Pattern Recognition (CVPR)},
  year={2018}
}

@inproceedings{scienceqa,
  title={Learn to Explain: Multimodal Reasoning via Thought Chains for Science Question Answering},
  author={Lu, Pan and Mishra, Swaroop and Xia, Tony and Qiu, Liang and Chang, Kai-Wei and Zhu, Song-Chun and Tafjord, Oyvind and Clark, Peter and Kalyan, Ashwin},
  booktitle={Advances in Neural Information Processing Systems (NeurIPS)},
  year={2022}
}

@inproceedings{llmpruner,
  title={{LLM-Pruner}: On the Structural Pruning of Large Language Models},
  author={Ma, Xinyin and Fang, Gongfan and Wang, Xinchao},
  booktitle={Advances in Neural Information Processing Systems (NeurIPS)},
  year={2023}
}

@article{lorap,
  title={{LoRAP}: Transformer Sub-Layers Deserve Differentiated Structured Compression for Large Language Models},
  author={Li, Guangyan and Zeng, Yongqiang and Huang, Jun and Zhang, Zhiqiang and Wang, Zhaocheng and Zhang, Xiaoxing and Zhao, Zhixuan},
  journal={arXiv preprint arXiv:2404.09695},
  year={2024}
}

@inproceedings{cfsp,
  title={{CFSP}: An Efficient Structured Pruning Framework for {LLM}s with Coarse-to-Fine Activation Information},
  author={Wang, Yuxin and Ma, Minghua and Wang, Zhang and Chen, Jingchang and Shan, Liping and Yang, Qing and Xu, Dongliang and Liu, Ming and Qin, Bing},
  booktitle={Annual Meeting of the Association for Computational Linguistics (ACL)},
  year={2025}
}
\bibliographystyle{conference}

\appendix

\section{Detailed analysis of power transformation exponent}
\label{sec:power_appendix}

This appendix provides comprehensive ablation experiments on the power transformation exponent $b$ to support our choice of $b=2.7$ in the main paper.

\subsection{Experimental setup}

We systematically evaluate different values of $b$ ranging from 1.0 to 3.0 on Llama-3-8B at 20\% pruning rate. For each value of $b$, we apply the full proposed method (power transformation + sign-preserving aggregation + outlier removal) and measure average accuracy across all five benchmark datasets (WinoGrande, HellaSwag, ARC-easy, ARC-challenge, MMLU).

\subsection{Results}

\begin{figure}[tb]
    \centering
    \includegraphics[width=0.9\columnwidth]{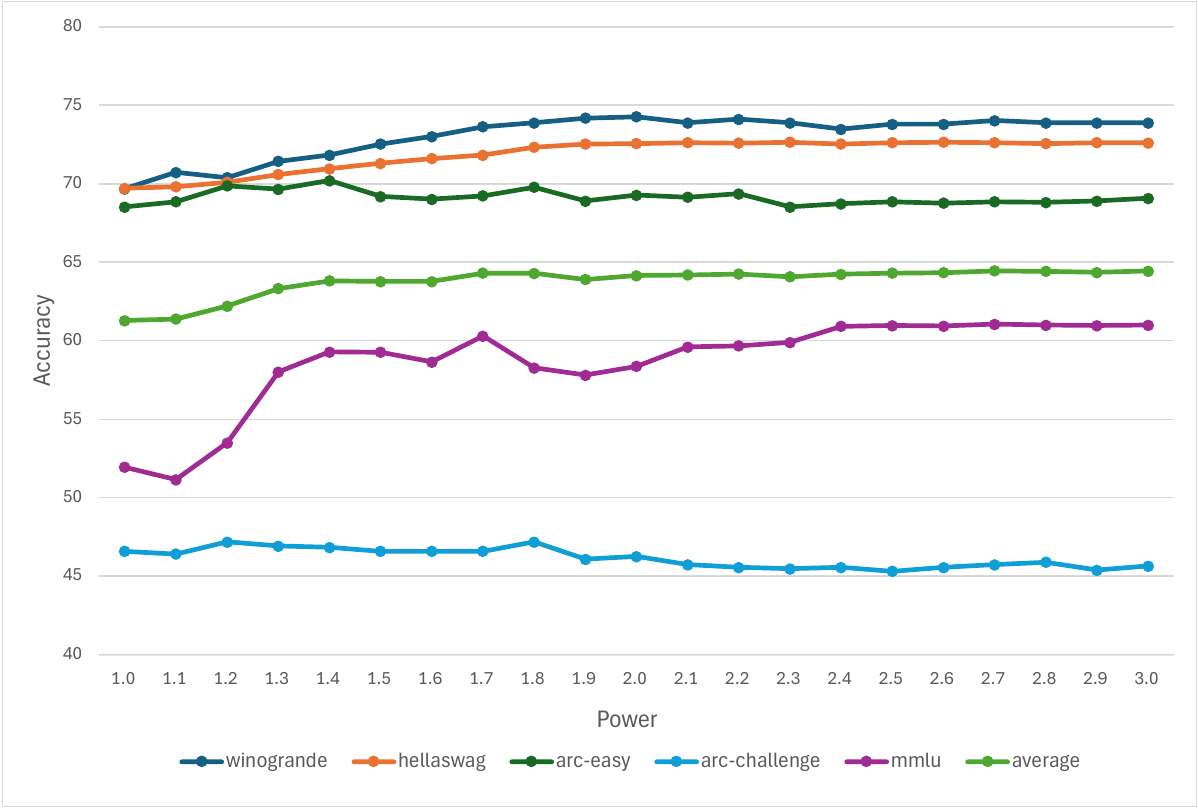}
    \caption{Ablation study of power transformation exponent $b$ on Llama-3-8B at 20\% pruning rate. The curve shows average accuracy across five benchmarks (WinoGrande, HellaSwag, ARC-easy, ARC-challenge, MMLU). Performance improves from 61.29\% at $b=1.0$ to a peak of 64.46\% at $b=2.7$, then plateaus in the range $b=2.5$--$3.0$ with minimal variation (0.15 percentage points).}
    \label{fig:power_analysis_appendix}
\end{figure}

Figure~\ref{fig:power_analysis_appendix} shows the relationship between the power transformation exponent $b$ and pruning performance. Table~\ref{tab:power_ablation} provides the detailed numerical results. The results demonstrate several key findings:

\textbf{Low exponent region ($b < 1.5$):} When $b$ is too small, the transformation provides insufficient noise suppression. At $b=1.0$ (no transformation), the method achieves 61.29\% average accuracy. Performance gradually improves as $b$ increases: 61.39\% at $b=1.1$, 62.21\% at $b=1.2$, reaching 63.82\% at $b=1.4$. In this region, the SNIP scores retain too much of their original densely-clustered distribution, and standardization amplifies noise from the low-importance region.

\textbf{Transition region ($b = 1.5$--$2.5$):} Performance continues to improve steadily, reaching 63.78\% at $b=1.5$ and peaking at 64.31\% at $b=1.7$. The curve shows some minor fluctuations but maintains an upward trend, with 64.15\% at $b=2.0$ and 64.31\% at $b=2.5$.

\textbf{Optimal region ($b = 2.5$--$3.0$):} Performance stabilizes at its highest level in this range. Starting from 64.31\% at $b=2.5$, accuracy reaches 64.35\% at $b=2.6$, peaks at 64.46\% at $b=2.7$, and maintains 64.43\% at $b=2.8$, 64.35\% at $b=2.9$, and 64.44\% at $b=3.0$. This plateau demonstrates remarkable robustness, with all values in the range of 64.31--64.46\% (a variation of only 0.15 percentage points). Within this range, the transformation optimally balances noise suppression and signal preservation.

\textbf{Performance gain:} Comparing the optimal value ($b=2.7$, 64.46\%) with no transformation ($b=1.0$, 61.29\%), the power transformation contributes a 3.17 percentage point improvement, demonstrating its significant impact on pruning quality.

\begin{table}[b]
\centering
\caption{Detailed ablation results for power transformation exponent $b$ on Llama-3-8B at 20\% pruning (accuracy \%).}
\label{tab:power_ablation}
\small
\begin{tabular}{@{}ccccccc@{}}
\toprule
$b$ & \textbf{WinoG} & \textbf{HellaS} & \textbf{ARC-e} & \textbf{ARC-c} & \textbf{MMLU} & \textbf{Avg} \\
\midrule
1.0 & 69.69 & 69.71 & 68.52 & 46.59 & 51.95 & 61.29 \\
1.5 & 72.53 & 71.30 & 69.19 & 46.59 & 59.27 & 63.78 \\
2.0 & 74.27 & 72.56 & 69.28 & 46.25 & 58.37 & 64.15 \\
2.5 & 73.80 & 72.62 & 68.86 & 45.31 & 60.97 & 64.31 \\
\textbf{2.7} & \textbf{74.03} & \textbf{72.62} & \textbf{68.86} & \textbf{45.73} & \textbf{61.05} & \textbf{64.46} \\
3.0 & 73.88 & 72.61 & 69.07 & 45.65 & 61.00 & 64.44 \\
\bottomrule
\end{tabular}
\end{table}

\subsection{Analysis of the performance curve}

The performance curve exhibits three distinct phases. In the initial phase ($b < 1.5$), the transformation is too weak, and performance improves monotonically as $b$ increases. The transition phase ($b = 1.5$--$2.5$) shows continued improvement with minor fluctuations, suggesting the transformation is approaching its optimal compression level. Finally, the plateau phase ($b \geq 2.5$) demonstrates that the method has reached its performance ceiling, where further increasing $b$ neither helps nor hurts significantly.

Interestingly, even at $b=3.0$, performance remains at 64.44\%, only 0.02 percentage points below the peak. This robustness is practically valuable, as it means practitioners do not need precise tuning of $b$. Any value in the range $[2.5, 3.0]$ will yield near-optimal results.

\subsection{Implications}

The existence of a clear optimal range with a relatively flat plateau has important practical implications. Users can apply $b=2.7$ without extensive hyperparameter tuning, and small deviations ($\pm 0.3$) have minimal impact on performance. This robustness makes the method practical for deployment across different models and datasets.

\end{document}